# Parallel development of social preferences in fish and machines

Joshua D. McGraw (jdmcgraw@iu.edu)[1,2], Donsuk Lee (donslee@iu.edu)[1,2], Justin N. Wood (woodjn@iu.edu)[1-4]

[1]Department of Informatics,
[2]Cognitive Science Program,
[3]Center for the Integrated Study of Animal Behavior,
[4]Department of Neuroscience,
Indiana University, Bloomington, IN 47408 USA

## Abstract

What are the computational foundations of social grouping? Traditional approaches to this question have focused on verbal reasoning or simple (low-dimensional) quantitative models. In the real world, however, social preferences emerge when high-dimensional learning systems (brains and bodies) interact with high-dimensional sensory inputs during an animal's embodied interactions with the world. A deep understanding of social grouping will therefore require embodied models that learn directly from sensory inputs using high-dimensional learning mechanisms. To this end, we built artificial neural networks (ANNs), embodied those ANNs in virtual fish bodies, and raised the artificial fish in virtual fish tanks that mimicked the rearing conditions of real fish. We then compared the social preferences that emerged in real fish versus artificial fish. We found that when artificial fish had two core learning mechanisms (reinforcement learning and curiosity-driven learning), artificial fish developed fish-like social preferences. Like real fish, the artificial fish spontaneously learned to prefer members of their own group over members of other groups. The artificial fish also spontaneously learned to self-segregate with their in-group, akin to self-segregation behavior seen in nature. Our results suggest that social grouping can emerge from three ingredients: (1) reinforcement learning, (2) intrinsic motivation, and (3) early social experiences with in-group members. This approach lays a foundation for reverse engineering animal-like social behavior with image-computable models, bridging the divide between high-dimensional sensory inputs and social preferences.

## Introduction

Social preferences are widespread across the animal kingdom. Individuals spontaneously organize into social groups, including bird blocks, fish shoals, insect swarms, and human crowds. Many animals, including humans, develop social preferences early in life, rapidly learning to favor "us" over "them" during social interactions (e.g., Al-Imari & Gerlai 2008; Engeszer et al. 2004; Kinzler et. al., 2007; Molenberghs, 2013). What are the origins of such social preferences? Despite significant interest in social preferences across psychology and neuroscience, we know relatively little about the core learning mechanisms that drive social preferences in human and nonhuman animals. What is the nature of the learning mechanisms—present in newborn animals—that cause social preferences to emerge so rapidly and flexibility?

It has not generally been possible to address this question because the field lacked experimental platforms for directly comparing the development of social preferences across newborn animals and computational models formalizing candidate learning algorithms. As a result, researchers could not test whether particular algorithms can actually learn animal-like social preferences. To evaluate whether a computational model learns like a newborn animal, we argue that an experimental platform must have two features. First, the animals and models must be raised in the same environments. This is essential because social preferences emerge both from the *learning algorithms* and the *training data* (experiences) acquired by the agent. Any observed differences in social preferences across animals and models could be due to differences in the algorithms, training data, or some combination of the two factors. Thus, evaluating whether computational models learn like animals requires 'raising' models in the same environments as newborn animals, giving the models and animals access to the same training data. Second, the animals and models must be tested with the same tasks. Psychologists have long recognized that preference behavior is task-dependent. To confirm that animals and models learned the same social preferences, the animals and models must be tested with the same tasks, thereby ensuring that any observed differences across the animals and models are not due to differences in the tasks themselves.

Here we introduce "digital twin" experiments that meet both requirements, allowing newborn animals and artificial animals (embodied learning algorithms) to be raised in the same environments and tested with the same tasks. Digital twin experiments involve first selecting a target animal study, and then creating digital twins (virtual replicas) of the animal environments in a video game engine. Artificial animals are then raised and tested in those virtual environments and their behavior is compared with the behavior of the real animals in the target study. By raising and testing real and artificial animals in the same environments, we can test whether animals and models spontaneously learn common social preferences.

As a starting point, we focused on the social preferences of newborn fish. We chose fish because they can be reared in controlled visual environments, are mobile on the first day after hatching, and rapidly learn social preferences based on visual information (Engeszer et al., 2004; Geng & Peterson, 2019; Hinz & de Polavieja, 2017; Ogi et al., 2021; Tallafuss & Bally-Cuif, 2003). For the target animal study, we focused on Engeszer et al. (2004). In the study, newly-hatched zebrafish (*Danio rerio*) were reared for several months in controlled visual environments that contained social partners (other fish) with one of two possible pigment types. Once the fish had been reared with social partners of one pigment type, the researchers used a two-alternative forced-choice (2AFC) task to test whether the fish had developed a preference for the familiar pigment type over the novel pigment type. The fish developed a strong preference for the familiar pigment type over the novel pigment type – independently of the fishes' own pigment type. This experiment thus reveals an important role of visual learning in the development of social preferences. To explore which learning algorithms can drive these early-emerging social preferences, we performed two experiments with artificial fish whose behavior was learned through intrinsically motivated reinforcement learning.

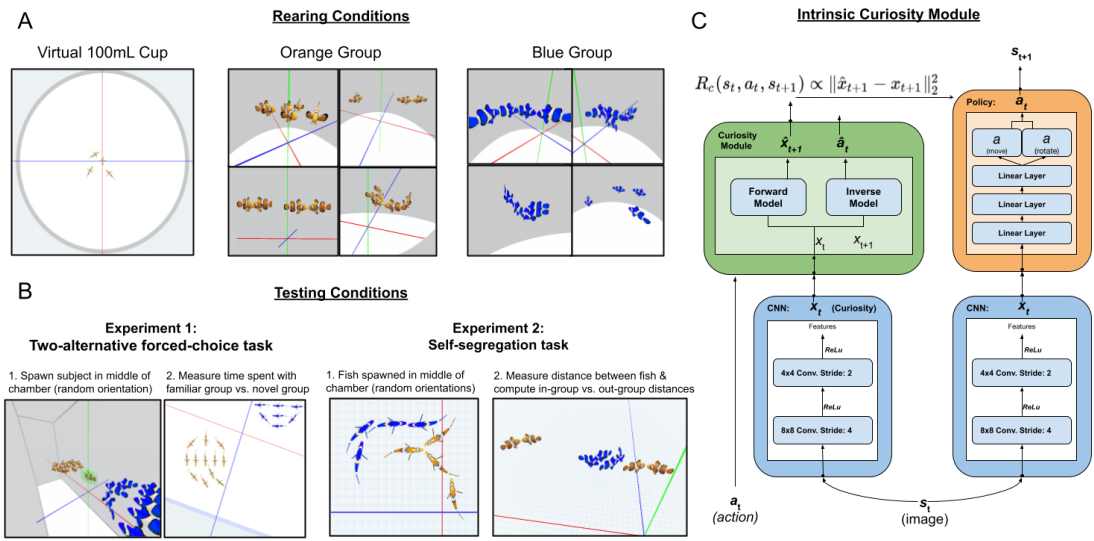

**Figure 1. Experimental Design. (A)** Rearing conditions of the artificial fish. Each group (orange & blue) was reared in a small white virtual cup—akin to the rearing conditions from Engeszer et al. (2004). **(B)** Testing conditions of the artificial fish. We tested the fish in two tasks. In Experiment 1, we used the 2AFC task from the target animal study (Engeszer et al., 2004). In Experiment 2, we used a self-segregation task that measured whether the fish spontaneously group into "us" versus "them" during social interactions. **(C)** The artificial neural network used in the artificial fish. The blue boxes denote visual encoders (CNNs), the green box denotes the curiosity module (intrinsic reward), and the orange box denotes the policy network used to select actions.

## Experiment 1: 2AFC Task

We first explored whether artificial fish learn similar social preferences as real fish when they are reared in the same environments and tested with the same task. To match the 2AFC testing conditions of the fish in Engeszer et al. (2004), we tested the artificial fish in virtual fish tanks that mimicked the real fish tank proportions (Figure 1).

To develop fish-like social preferences, the artificial fish needed to spontaneously learn to prefer other social agents, in the absence of explicit rewards or supervision. The artificial fish also needed to learn how to move through space, developing knowledge of their location and direction (ego-motion) so that they could navigate to their preferred social group. None of these abilities (ego-motion nor social preferences) were hardcoded into the artificial fish.

## Methods

Using a video game engine (Unity), we raised (trained) artificial fish in realistic virtual environments akin to the fish tanks described in Engeszer et al. (2004). Due to its flexibility and power, Unity is an ideal testbed for AI simulation. In particular, Unity's development team actively supports a package known as 'ML-Agents Toolkit' (Juliani et al., 2018), which allows researchers to train artificial agents in virtual worlds. We used the *Unity ML-Agents* Package version 2.0.1 with Python 3.8.10 and the Python *mlagents* library version 0.26.0.

**Artificial Fish.** Real fish learn from raw sensory inputs and perform actions in 3D environments, driven by self-supervised learning objectives. Thus, to directly compare the real and artificial fish, we used 'pixels-to-actions' artificial fish that learn from raw sensory inputs and perform actions in 3D environments, driven by self-supervised learning objectives (intrinsic motivation).

We generated the artificial fish by embodying self-supervised learning algorithms in virtual animated fish bodies. The fish bodies measured 1.2 units (length) by 0.7 units (height) and the fish received visual input through an invisible forward-facing camera attached to its head (64×64 pixel resolution). The artificial fish could move themselves around the 3D environment by moving either forward, left, or right on every step. The plane of motion was restricted to a single flat plane in order to mimic the action space of a thin, shallow, layer of water commonly used in zebrafish research to prevent motion along the height axis.

To construct the artificial fish brains (Figure 1C), we used two biologically-inspired learning mechanisms: (a) deep reinforcement learning and (b) curiosity-driven learning. Deep reinforcement learning provides a computational framework for learning adaptive behaviors from high-dimensional sensory inputs. In reinforcement learning, agents maximize their long-term rewards by performing actions in response to their environment and internal state. To succeed in environments with real-world complexity, agents must learn abstract and generalizable features to represent the environment. Deep reinforcement learning combines reinforcement learning with deep neural networks in order to transform raw sensory inputs into efficient representations that support adaptive behavior. Artificial agents trained through deep reinforcement learning can develop human and animal abilities. For example, artificial agents can learn to play simple and complex video games (e.g., Atari: Mnih et al., 2015; Quake III: Jaderberg et al., 2019), develop advanced motor behaviors (e.g., walking, running, and jumping: Haarnaja et al., 2019), and learn to navigate 3D environments (Banino et al., 2018). We used a standard reinforcement learning algorithm called Proximal Policy Optimization (PPO) (Schulman et al., 2017).

The second mechanism—curiosity-driven learning—can drive the development of complex behaviors through self-supervised learning objectives (e.g., Haber et al., 2018; Oudeyer & Smith, 2016; Schmidhuber, 2010). Curiosity is a popular approach for endowing artificial agents with intrinsic motivation and involves rewarding agents based on how surprised they are by the world. Curiosity promotes learning by motivating agents to seek out informative experiences. By seeking out less predictable (and more informative) experiences, agents can gradually expand their knowledge of the world, continuously acquiring useful experiences for improving perception and cognition. This type of self-supervised learning is thought to resemble learning in real animals, who often learn about the world not by pursuing any specific goal, but rather by playing and exploring for the novelty of the experience (Gopnik et al., 2017).

We implemented curiosity-driven learning using a popular algorithm called Intrinsic Curiosity Module (ICM) (Pathak et al., 2017; Burda et al., 2018). The ICM module contains two separate neural networks: a forward and an inverse model. The inverse model is trained to take the current and next visual observation received by the agent, encode both observations using a single encoder, and use the result to predict the action that was taken between the two observations. The forward model is then trained to take the encoded current observation and action to predict the encoded next observation. The difference between the predicted and real encodings is then used as the intrinsic reward, and fed to the PPO algorithm. Bigger differences mean bigger surprise, which in turn means bigger intrinsic reward. By linking these two models together, the reward captures not only surprising things, but surprising things that the agent has control over (based on the agent's actions). This artificial curiosity allows machines to be trained without any extrinsic rewards from the environment, with learning driven entirely by intrinsic motivation signals.

We used an off-the-shelf ICM algorithm implemented in ML-Agents. ML-Agents allows artificial agents to be configured according to several hyperparameters, including the learning policy, the learning rate, and other common neural network configuration settings (e.g. batch size, layer width, etc.). We used the hyperparameters listed in Table 1. All of the artificial fish had the same network architecture: 2-layer CNN connected to a multilayer perceptron (see Table 1 for hyperparameters).

| Table 1: ML Agents Trainer configuration. ||
|---|---|
| **trainer_type: ppo** | **reward_signals: curiosity** |
| num_layers: 3 | num_layers: 3 |
| hidden_units: 512 | hidden_units: 128 |
| learning_rate: 0.0003 | learning_rate: 0.0003 |
| batch_size: 256 | strength: 1.0 |
| buffer_size: 2048 | gamma: 0.99 |
| beta: 0.01 | vis_encode_type: simple (2-layer CNN) |
| epsilon: 0.2 | normalize: false |
| lambd: 0.95 | max_steps: 1000000 |
| learning_rate_schedule: linear | time_horizon: 128 |

**Rearing Conditions.** To simulate the rearing conditions of real fish, we created two groups of differently-pigmented artificial fish: four blue fish and four orange fish. The orange fish were reared together in one group, and the blue fish were reared together in a separate group. Each group was reared in a small white virtual cup—akin to the rearing conditions from Engeszer et al. (2004).

During training (Figure 1A), the artificial fish experienced 1,000 training episodes. At the beginning of each episode, the fish were spawned at the center of the environment and rotated in a random direction. Each episode lasted for 1,000 actions ('steps') in the environment. The artificial fish were provided with a 140° field of view and could take one full action on every frame of the simulation. One full action was the result of two discrete movement sets: (a) [forward or stay] and (b) [rotate left, rotate right, or no rotation]. For example, an agent might decide to move [forward] + [left] on one frame, and then move [forward] + [no rotation] on the next frame. A sharp right turn would then be the result of [stay] + [rotate right] for several frames. Each rotation was ~2 degrees of rotation along the Y-axis per step.

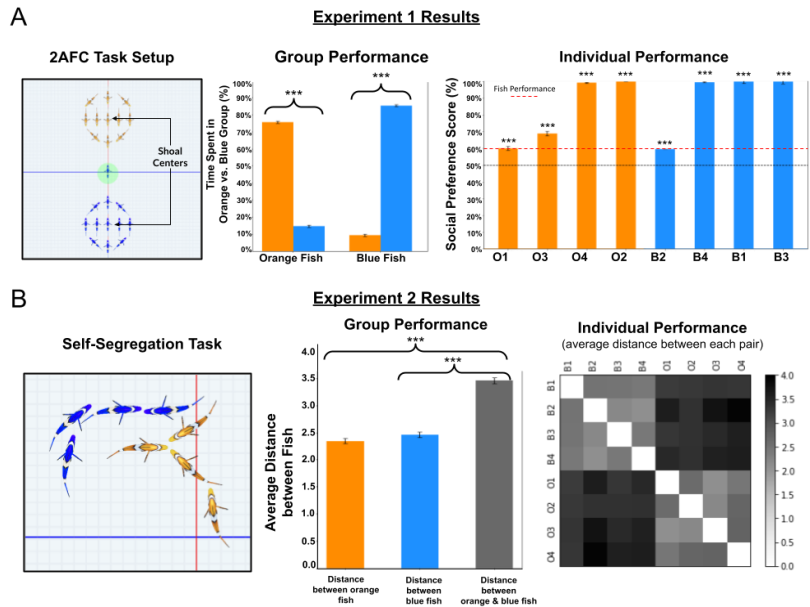

**Figure 2. (A)** Results on the 2AFC task. On the group level, the fish reared in the orange group spent more time with orange fish versus blue fish, and the fish reared in the blue group spent more time with blue fish versus orange fish. On the individual level, all of the artificial fish developed statistically significant social preferences for in-group members. The black dashed line denotes chance performance, and the red line denotes the fish performance from Engeszer et al. (2004). **(B)** Results on the self-segregation task. On the group level, the fish reared in the orange group self-segregated with orange fish versus blue fish, and the fish reared in the blue group self-segregated with blue fish versus orange fish. On the individual level, all eight of the artificial fish were more likely to self-segregate with in-group members. Lighter colors indicate smaller distances between fish. Significance levels are indicated: (*p ≤ .05, **p ≤ .01, ***p ≤ .001).

All of the artificial fish had the same learning algorithms, hyperparameters, and network architectures. However, each artificial fish started with a different random initialization of connection weights, and each fish's connection weights were shaped by its own particular experiences during the training phase. The artificial fish were trained for 1 million time steps, using a rack of eight NVIDIA A10 GPUs on a single Linux server environment.

**Testing Conditions.** After the training phase, the brain weights of the artificial fish were frozen for the test phase (i.e., the algorithms did not receive any rewards during the test phase and learning was discontinued). To mimic the testing conditions of the real fish, we tested the artificial fish using a 2AFC task (Figure 1B). Each of the eight artificial fish agents were tested separately across 1,000 test trials. At the start of each test trial, the test fish was placed in the center of the chamber in a random orientation. The chamber contained two shoaling groups (n = 11 fish): one group had a familiar pigment type (i.e., "orange" for fish reared in the orange group and "blue" for fish reared in the blue group), while the other group had a novel pigment type (i.e., "orange" for fish reared in the blue group and "blue" for fish reared in the orange group).

The fish in the two shoals generated the same swimming motion as the fish in the training phase. However, the shoaling fish remained in a stationary configuration across the test trials to prevent the movements of the shoaling fish from influencing the behavior of the focal (test) fish. As a result, the behavior of the fish in the blue and orange groups were identical. Each test trial consisted of 3,000 actions ('steps'). At every time step, we recorded the position of the test fish in (X, Y) coordinates. As with the real fish, we measured the proportion of time that the artificial fish spent in proximity to the group with the familiar pigment type versus the novel pigment type (measured as the distance to the center of the shoal).

## Results & Discussion

**Group performance.** Figure 2A shows the social preferences of the artificial fish. On the group level, the artificial fish spent significantly more time near the group with the familiar pigment type versus the novel pigment type ($t(7) = 5.27$, $p < .00001$). On average, the group spent 85.9% (SEM = 8%) of their time with in-group versus out-group fish (chance = 50%), indicating that artificial fish can develop fish-like social preferences when reared in similar environments as real fish.

**Individual-subject performance.** Since we collected a large number of test trials from each artificial fish, we also explored whether individual differences emerged across the fish. To test for the presence of individual differences, we examined whether the identity of the subject was a predictor of social grouping behavior. A one-way ANOVA showed that the identity of the subject was a strong predictor of performance: $F(7, 999) = 8.923$, $p < .00001$. All eight of the artificial fish showed a statistically significant preference for the in-group versus the out-group (all $P$s < .00001). Five of the artificial fish developed a strong preference for in-group members, spending more than 90% of their time with the familiar group. The other three fish developed less strong social preferences, spending 55% to 63% of their time with in-group members. Despite being trained in identical visual environments, the artificial fish developed different social behaviors as one another.

## Experiment 2: Self-Segregation Task

A defining signature of social preferences in the real world is that they drive animals to self-segregate into social groups. To test whether our artificial fish developed this signature of social grouping, we created a self-segregation task that involved placing all of the artificial fish in the same environment and measuring whether the fish spontaneously self-segregate into groups based on their pigment type. To be clear, Exp. 2 was not a digital twin experiment of a prior animal study, but rather a validity check that the 2AFC task used in Exp. 1 reliably captures the core construct under investigation: social grouping. If so, then the artificial fish from Exp. 1 should self-segregate into groups.

**Rearing Conditions.** The rearing conditions were identical to those used in Experiment 1.

**Testing Conditions.** The self-segregation task (Figure 1B) involved placing all of the 8 trained artificial fish (4 orange fish and 4 blue fish) into a single environment. At the start of each trial, the fish were centered in the environment, oriented randomly, and then allowed to freely move and interact with other fish. To measure whether the fish self-segregated, we measured the Euclidean distance between each fish and every other fish at each time step, then computed the in-group distance (i.e., the average distance to fish of the same color) and the out-group distance (i.e., the average distance to fish of unfamiliar color). We tested the fish across 1,000 trials. Each trial lasted 3,000 time steps.

## Results & Discussion

**Group performance.** Figure 2B shows the self-segregation behavior of the artificial fish. On the group level, the artificial fish spent significantly more time near fish with familiar versus novel colors, $t(7) = 15.5$, $p < .0001$. The distance to in-group members was significantly smaller than the distance to out-group members, indicating that the artificial fish spontaneously learned to self-segregate into social groups.

**Individual-subject performance.** As illustrated in Figure 2B, all eight of the artificial fish showed a statistically significant preference for in-group versus out-group members (all $P$s < .00001). To test for the presence of individual differences, we examined whether the identity of the subject was a predictor of self-segregation behavior. A one-way ANOVA showed that the identity of the subject was a strong predictor of performance: $F(7, 999) = 720.5$, $p < .00001$. These results extend the results from Experiment 1, showing that artificial fish can spontaneously develop self-segregation behavior, favoring "us" over "them."

## Discussion

We performed digital twin experiments, in which newborn fish and artificial fish were raised and tested in the same visual environments. This approach permits direct comparison of whether animals and machines learn common social preferences when exposed to the same experiences (training data). In this paper, we explored whether artificial fish can spontaneously learn fish-like social preferences, leading to social grouping and a preference for "us" over "them." We found that fish-like social grouping spontaneously develops in artificial fish who are equipped with reinforcement learning and curiosity-driven learning. These social preferences emerged when artificial fish were reared in similar visual (and social) environments as real fish (Engeszer et al., 2004).

**Origins of social grouping.** Biologists and psychologists have proposed a range of theories about the origins and computational foundations of social behavior (e.g., Pinker, 1994; Cosmides et al., 2003; Tinbergen, 1951). Some theorists have argued that social behavior emerges from innate, domain-specific learning mechanisms for representing social partners and for categorizing the social world into groups (e.g., Spelke & Kinzler, 2007), whereas others have argued that social behavior emerges from domain-general learning mechanisms that become specialized for social cognition during development (e.g., Oudeyer & Smith, 2016). It has not been possible to distinguish between these theories because researchers could not test whether candidate learning mechanisms are sufficient to produce animal-like social behavior. Do embodied agents actually need innate, domain-specific mechanisms to learn about the social world? Or can social knowledge emerge from domain-general mechanisms shaped by embodied social experiences?

Our experiments provide an existence proof that social grouping can emerge from domain-general mechanisms. Neither reinforcement learning nor curiosity-driven learning were designed to produce social grouping in embodied agents. Nevertheless, when these learning algorithms are embodied and trained in realistic social environments, animal-like social preferences spontaneously develop. Thus, we hypothesize that social grouping is an *emergent property* of three ingredients: reinforcement learning, intrinsic motivation (e.g., curiosity-driven learning), and embodied social experiences with in-group members.

Consequently, evolution would not have needed to discover innate, domain-specific mechanisms to produce social behavior. Rather, once evolution discovered domain-general mechanisms, animals could have rapidly learned social behavior during early development. Thus, we speculate that the computational foundations of social behavior are *domain-general learning mechanisms* (e.g., reinforcement learning and intrinsic motivation), which allow animals to rapidly learn domain-specific social knowledge through early social experiences.

Why do social preferences emerge from domain-general learning mechanisms? Perhaps counterintuitively, we found that curiosity (i.e., a preference for the unpredictable) during early stages of life drove social grouping (i.e., a preference for familiar groupmates). This preference for familiar groupmates developed in the artificial fish because the other fish were the most unpredictable things in the environment. Accordingly, curiosity-driven learning motivates agents to learn about groupmates, leading to the development of collective behavior (Lee, Wood, & Wood, 2021) and social grouping. We emphasize that curiosity-driven learning must be subject to a learning window to produce social grouping; if learning remains permanently "active," then agents will continue to seek unpredictable experiences and will likely not develop a preference for in-group members. Critically, there is ample evidence for learning windows in nature. For example, filial imprinting occurs during a short sensitive period immediately after birth, during which animals develop a lasting attachment to groupmates seen early in life (Horn, 2004; Patterson & Bigler, 2006; Tinbergen, 1951). We speculate that reverse engineering animal behavior and intelligence will require not only finding the right learning mechanisms, but will also require raising animals and machines in the same environments and constraining machine learning to similar learning windows as real animals.

**Image computable models of social grouping.** Lee et al. (2021) introduced pixels-to-actions models of collective behavior that spontaneously develop animal-like grouping behavior. Our study extends this approach to the domain of social preferences. Pixels-to-actions models are valuable because they formalize the mechanisms underlying the entire learning process, from sensory inputs to behavioral outputs. As a result, these image computable models can be directly compared against real animals, falsified, and refined.

These results set the stage for many exciting future directions. With the discovery of a model class that spontaneously learns similar social preferences as real animals, we can now search through the model class to find particularly strong models, via a continuous cycle of model creation, model prediction, and model testing against experimental results. This approach can move the field forward by encouraging the creation of neurally mechanistic pixels-to-actions models that learn from the same embodied data streams as newborn animals. Over time, we can cull models that are less accurate and focus attention on improving and extending the most accurate models.

The most promising models can then be investigated in richer detail, leading to greater intuitive understanding of the underlying mechanisms. Controlled comparisons with different architectures, objective functions, and learning rules could define the necessary and sufficient learning algorithms for animal-like social behavior. Likewise, controlled comparisons using the same learning algorithms, but different training data (from different controlled-rearing experiments), could reveal which experiences are necessary and sufficient to develop animal-like social behavior.

Finally, pixels-to-actions models are valuable because they allow integration of research findings across diverse fields. For instance, while it was not our primary goal, we observed robust evidence for 'personality differences' in the artificial fish, with some fish strongly preferring to spend most of their time with groupmates and other fish developing less strong social preferences. We emphasize that the artificial fish had identical learning algorithms and were raised in the same environment. The only difference between the fish was the initial random configuration of their brain weights and the particular experiences that the fish acquired during the training phase. Nevertheless, these differences were sufficient to produce personality differences in machines. We suspect that the complex group dynamics that emerged during the training phase resulted in different learning opportunities (training data) for each fish, which led each fish to learn different policies (personalities) from one another. Future research could explore this idea directly by characterizing how the particular experiences encountered by artificial fish lead to various personality differences (e.g., sociality, boldness, novelty seeking).

Importantly, pixels-to-actions models allow researchers to study different phenomena (i.e., social preferences and personality differences) in parallel, using the same integrative pixels-to-actions model. Future studies might further expand the reach of these integrative models, by extending this approach to other collective behavior topics, including leadership, foraging, and navigation.

## Conclusion

In sum, we present a pixels-to-actions model of social preferences, which indicates that we have isolated a set of learning mechanisms that are sufficient for learning social preferences in embodied agents. When trained in naturalistic environments, domain-general learning mechanisms are sufficient to drive social grouping. We thus see these results as a starting point for building an engineering-level understanding of the mechanisms that underlie social behavior. Our results complement a growing body of work using deep neural networks to model the visual (Yamins et al., 2014), auditory (Kell et al., 2018), and motor (Michaels et al., 2020) systems, and extend this "reverse engineering" approach to the study of social grouping: a behavior with deep historical roots in psychology, neuroscience, and biology.

## Acknowledgments
Funded by NSF CAREER Grant BCS-1351892 and a James S. McDonnell Foundation Understanding Human Cognition Scholar Award.